\pdfoutput=1

\documentclass[11pt]{article}

\usepackage{acl}

\usepackage{times}
\usepackage{latexsym}
\usepackage{multirow}
\usepackage{amsmath}
\usepackage{amssymb}

\usepackage[T1]{fontenc}

\usepackage[utf8]{inputenc}
\usepackage{graphicx}
\usepackage{microtype}
\usepackage{tcolorbox}
\usepackage{inconsolata}

\usepackage{algorithm}
\usepackage{amsmath}
\usepackage{algpseudocode}

\usepackage{enumitem}

\title{Multi-Perspective Consistency Enhances Confidence Estimation in Large Language Models}
\author{Pei Wang$^{1*}$, Yejie Wang$^{1*}$, Muxi Diao$^{1*}$, Keqing He$^{2}$, Guanting Dong$^{1}$, {\bf Weiran Xu$^{1}$} \thanks{\ The first three authors contribute equally. Weiran Xu is the corresponding author.}\\
  $^1$Beijing University of Posts and Telecommunications, Beijing, China\\
$^{2}$Meituan, Beijing, China\\
  \texttt{\{wangpei,wangyejie,dmx,dongguanting,xuweiran\}@bupt.edu.cn}\\
  \texttt{\{hekeqing\}@meituan.com}
}

\begin{document}
\maketitle
\begin{abstract}
In the deployment of large language models (LLMs), accurate confidence estimation is critical for assessing the credibility of model predictions. However, existing methods often fail to overcome the issue of overconfidence on incorrect answers. In this work, we focus on improving the confidence estimation of large language models. Considering the fragility of self-awareness in language models, we introduce a Multi-Perspective Consistency (\textbf{MPC}) method. We leverage complementary insights from different perspectives within models (\textbf{MPC-Internal}) and across different models (\textbf{MPC-Across}) to mitigate the issue of overconfidence arising from a singular viewpoint. The experimental results on eight publicly available datasets show that our MPC achieves state-of-the-art performance. Further analyses indicate that MPC can mitigate the problem of overconfidence and is effectively scalable to other models. \footnote{We will open-source our code and all the evaluation results to facilitate future explorations.}
\end{abstract}

\section{Introduction}
Large language models, such as GPT-4 \cite{openai2023gpt4}, have achieved outstanding performance in multiple downstream NLP tasks \cite{sanh2021multitask,chung2022scaling,yuan2023scaling,luo2023wizardcoder,dong2023abilities}. However, as models are increasingly used in practical applications, it is important to accurately assess their confidence \cite{guo2017calibration,tomani2021trustworthy}. Confidence estimation is to evaluate the uncertainty of the model prediction, which is critical for ensuring the clarity and trustworthiness of human-machine interaction \cite{kuleshov2018accurate,xiao2021hallucination,kuleshov2022calibrated,song2023large}. 

The standard approach of estimating confidence is to use the softmax probabilities of these models. However, due to the unavailability of the logits, as the most powerful LLM currently available is closed-source, researchers employ two alternative methods for confidence estimation. one is Verbalized-based method (\textbf{Verb}) \cite{kadavath2022language,lin2022teaching,tian2023just}. They prompt LLMs to provide a confidence probability verbally and optimize the prompt template by combining techniques like CoT \cite{xiong2023llms} and TOT \cite{yao2023tree}. The other one is Self-Consistency Confidence (SC) \cite{wang2023selfconsistency,xiong2023llms}, which calculates the probability of the answer appearing as the confidence. The essence is to measure the correctness of the answer by using the consistency between answers.

\begin{figure}[t]
    \centering
    \resizebox{1.0\linewidth}{!}{
	\begin{minipage}{0.45\linewidth}
		\vspace{3pt}
		\centerline{\includegraphics[width=\textwidth]{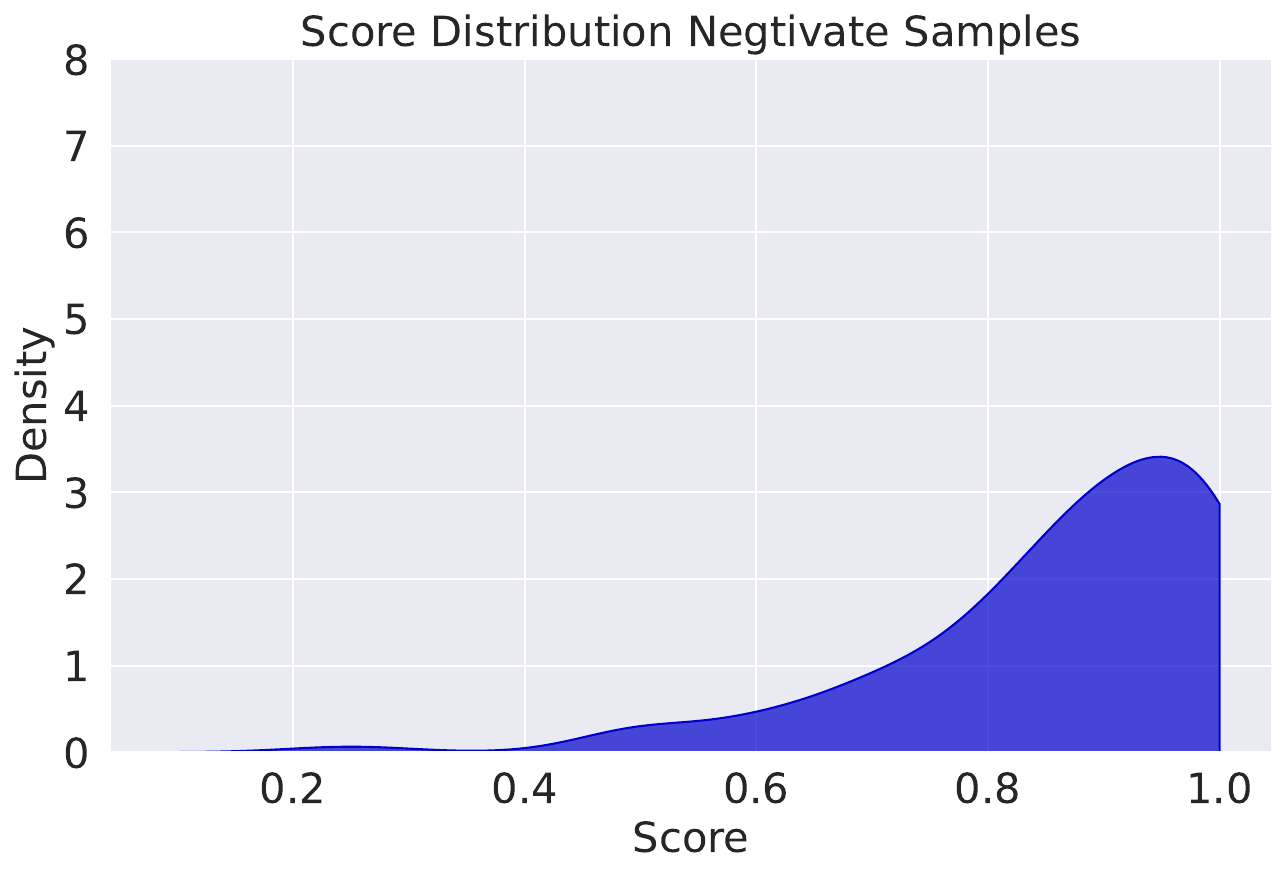}}
		\centerline{(a) Verb} 
	\end{minipage}
 
	\begin{minipage}{0.45\linewidth}
		\vspace{3pt}
		\centerline{\includegraphics[width=\textwidth]{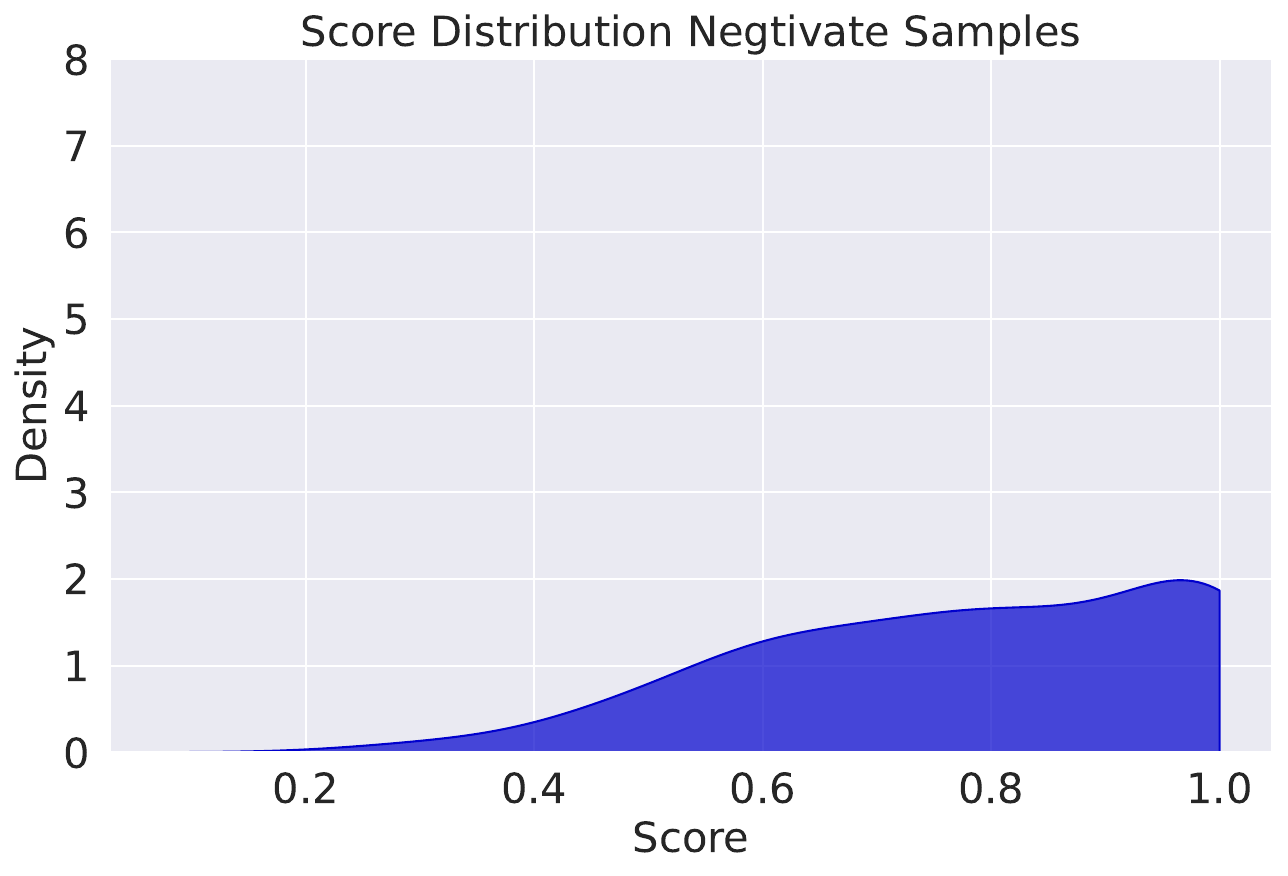}} 
		\centerline{(b) SC}
	\end{minipage}
    }
	\caption{Confidence score distribution of GPT-4's incorrect samples in TruthfulQA. The horizontal axis represents the confidence scores predicted under this method, and the vertical axis represents the probability density. Theoretically, it is preferable for the entire distribution to shift \textbf{left} as far as possible.}
	\label{fig:intro}
\end{figure}

\begin{figure*}
    \centering
    \resizebox{1.0\linewidth}{!}{
    \includegraphics{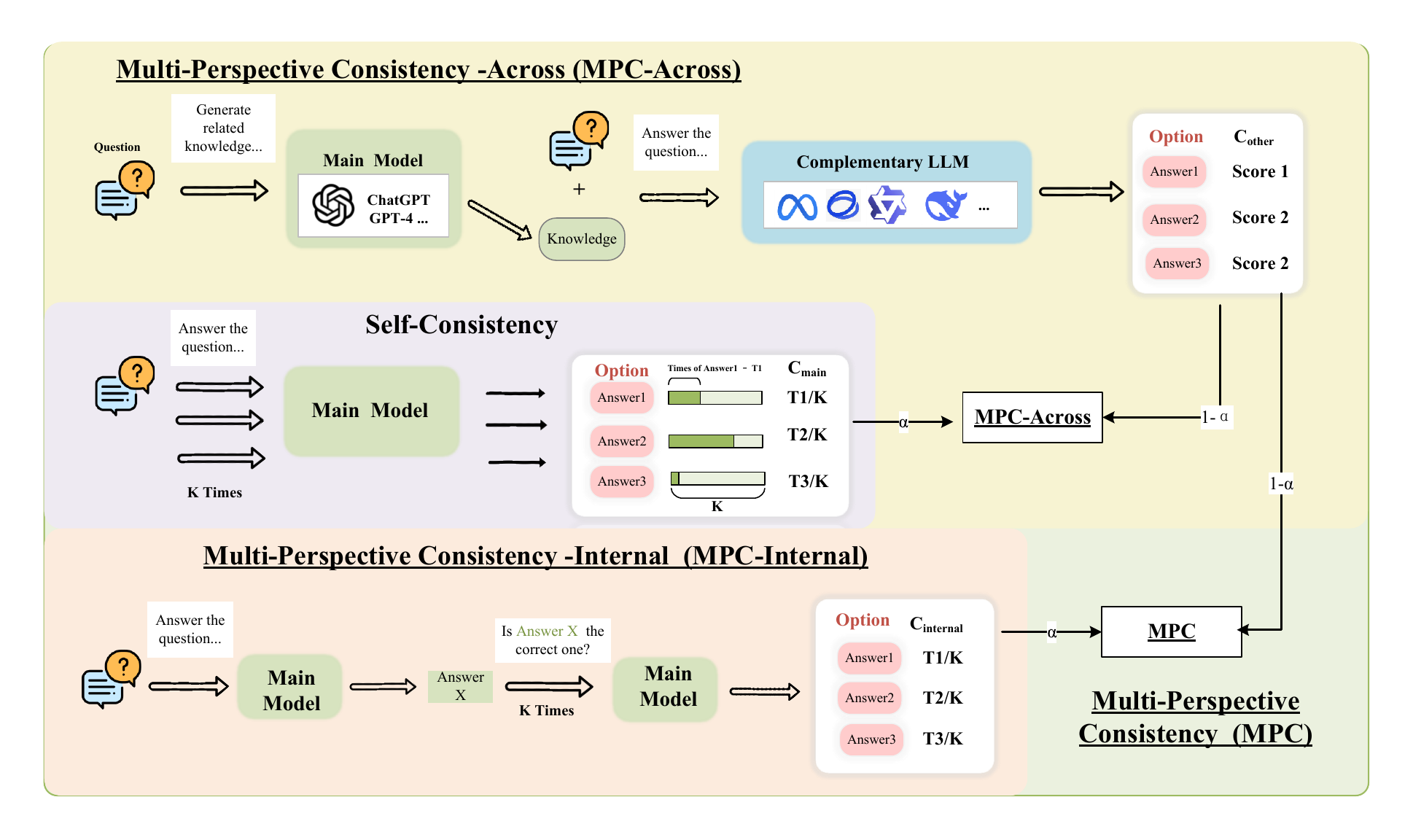}}
    \caption{The overall architecture of our Complementary  Perspective Consistency for confidence estimation.}
    \label{overall_structure}
    \vspace{-0.45cm}
\end{figure*}
\renewcommand{\dblfloatpagefraction}{.8}

However, these works bring a common limitation where LLMs demonstrate a significant level of overconfidence even when they provide incorrect answers \cite{xiong2023llms,tian2023just,shrivastava2023llamas}. This raises the question of whether current instruction-following models can truly recognize their own errors. Figure \ref{fig:intro} illustrates the confidence distribution of incorrect answers under different confidence estimation paradigms, revealing that these approach exhibits a more pronounced issue of overconfidence and only a few instances of incorrect answers can be assigned a lower confidence score.

In this work, we propose a \textbf{M}ulti-\textbf{P}erspective \textbf{C}onsistency (\textbf{MPC}) method. Considering the fragility of self-awareness in language models, MPC leverages complementary insights from different perspectives to mitigate the issue of overconfidence arising from a singular viewpoint. Specifically, MPC obtains confidence scores from multiple perspectives through MPC-Internal and MPC-Across to achieve better confidence estimation by fusing them. MPC-internal prompts LLMs to reconsider the questions from the verifier's perspective. We find that MPC-internal mitigates the overconfidence of LLMs when the generated answers are inconsistent under two different perspectives. Compared to MPC-Internal, MPC-Across utilizes stronger perturbations by considering answers generated by different models. We collect the confidences of different models and obtain the MPC-Across confidence through weighted averaging.
Our experiments demonstrate that both MPC-Internal and MPC-Across yield the SOTA performance on existing estimation methods, and combining them together can produce more robust confidence estimation results. Further analyses indicate that MPC can mitigate the problem of overconfidence and exhibit good generality.

Our contributions are: 
\begin{itemize}
  \item We are the first to propose the use of Multi-Perspective to alleviate the problem of overconfidence in confidence estimation. 

  \item We introduce two methods, MPC-Internal and MPC-Across, which mitigate overconfidence by incorporating internal self-validation and cross-model perspective integration. 

  \item 
  We conduct extensive experiments on eight publicly available datasets, and the results show that MPC exceeds the existing strong baselines. Further experimental analysis shows that MPC can alleviate overconfidence issues to some extent and be easily extended to other models.
\end{itemize}

\section{Methodology} 
In this section, we elaborate on the specifics of our methodology. As shown in Figure \ref{overall_structure}, we will introduce our method from two dimensions, \textbf{MPC-Internal} and \textbf{MPC-Across}, which alleviate the issue of overconfidence from different directions.

\subsection{Problem Formulation}
Confidence estimation is employed to assess the uncertainty of model predictions. For any given input $X$ and its corresponding model prediction $\hat{y} = f(x)$, we aim to get a function $ g: \mathcal{X} \times \mathcal{Y} \rightarrow [0, 1] $ that outputs a confidence score $c$, quantifying the reliability of the prediction $ \hat{y} $. The objective is to ensure that $c$ accurately reflects the probability of prediction correctness.

\subsection{MPC-Internal}

MPC-Internal introduces an internal verification mechanism to the model when generating answers, mitigating the overconfidence in SC. Specifically,
As shown in the bottom of Figure \ref{overall_structure}, in each inquiry, instead of directly asking the model to provide a definitive answer, we ask the model to analyze the correctness of a specific option and then make a final judgment based on its analysis. The related prompt is introduced in Appendix \ref{appendix:prompt_psc}.

For each question, we generate multiple answers and then use the following formula to calculate the confidence estimate: $C_{i} = \frac{T_{i}}{K}$. Here, $C_{i}$ represents the model's confidence in $i^{th}$ answer under MPC-internal; $T_{i}$ is the times of  $i^{th}$ answers; and $K$ is the total number of answers. In keeping with previous studies, we set $K = 15$. In Section \ref{sec:k}, we conduct ablation experiment on K.

\subsection{MPC-Across}
MPC-Across leverages the varying reasoning abilities of different models. After extensive pre-training on large amounts of language data, Large Language Models have demonstrated a strong reservoir of knowledge and reasoning abilities. However, due to the variations in model size, training parameters, and training corpora among different models, they exhibit differences in their performance at a detailed level. We argue that the differing reasoning abilities of different models can provide Multi-Perspectives to alleviate the overconfidence issue of a single model in wrong answers. Based on that point we propose MPC-Across which includes the following key steps:

\textit{1.} For the main model, we sample multiple answers for each problem and use the frequency of each answer as its initial confidence estimation $C_{main}$ same with SC. We choose the answer with the highest frequency to calculate the metric. It's shown at the middle of Figure \ref{overall_structure}.

\textit{2.} We use the self-consistency or logits-based method to collect scores $C_{other}$ from different models for the question as shown in the top of Figure \ref{overall_structure}. To alleviate the negative impact of poor fusion ability of smaller models, we prompt main models to generate explanations and offer to other models. 

\textit{3}. Weighted average will be used to fuse the confidence estimates from different models, where the main model's estimate is given a weight parameter $\alpha$, and the other one is weighted by $1-\alpha$. We select $\alpha=0.8$. 

$ C_{across} = \alpha \cdot C_{main} + (1 - \alpha) \cdot C_{other} $

In Section \ref{sec:alpha}, we conduct ablation experiment on the value of $\alpha$ to demonstrate that MPC-Acorss is robust to $\ alpha $.

\subsection{MPC}

MPC integrates both MPC-Internal and MPC-Across approaches to take full advantage of both. Specifically, When conducting confidence estimation with MPC-Across, we initially use the MPC-Internal method as a substitute for SC to perform the preliminary evaluation. The final confidence score is given by the formula:

$ C_{MPC} = \alpha \cdot C_{internal} + (1 - \alpha) \cdot C_{other} $

We summarize the pseudo-code of Perturbed-Consistency in Algorithm \ref{algorithm}.

\begin{algorithm*}
\caption{Multi-Perspective Consistency}
\label{algorithm}
\begin{algorithmic}[1]
\Require {Question $Q$, Options $O$ from 1 to $m$, Main model $M_{m}$, Complementary Model $M_{other}$, Number of inquiries $K = 15$, Weight parameter $\alpha = 0.8$ }
\Ensure Confidence score of Q $C_{final}$ 
\State Initialize count $T_i \gets 0$ for each answer option $Oi$
\State Prompt $M_{m}$ to generate a Answer $Ans$ for $Q$
\For{$k = 1$ to $K$}  \Comment{\textbf{Multi-Perspective Consistency-Internal}}
    \State Ask $M_{m}$ to analyze $Ans$ and provide an answer $A_k$
    \State Increment count $T_{A_k} \gets T_{A_k} + 1$
\EndFor
\For{each answer option $i$}
    \State Calculate confidence Score $C_{internal_i} = \frac{T_i}{K}$
\EndFor
\State Prompt $M_{m}$ generate related knowledge $K*$ for Q \Comment{\textbf{Multi-Perspective Consistency-Across}}
\State Generate a prompt using $Q$ and $K*$
\State Use $M_{other}$ and designed prompt to infer Q
\For{$i = 1$ to $m$}
    \State Obtain $C_{other_i}$ of the answer option $O_i$ by $M_{other}$
\EndFor
\For{each answer option $i$}  \Comment{\textbf{Multi-Perspective Consistency}}
\State Calculate final confidence $C_{MPC} = \alpha \cdot C_{internal_i} + (1 - \alpha) \cdot C_{other_i}$ 
\EndFor
\end{algorithmic}
\end{algorithm*}

\section{Experiments}

\subsection{Experimental Settings}

\textbf{Dataset}
We utilize eight commonly used public datasets for evaluation, including four domain subsets of MMLU \cite{hendrycks2021measuring}, Chemistry, Computer\_Security, Business\_Ethics and Anatomy. Other datasets include TruthfulQA \cite{lin2022truthfulqa}, CSQA \cite{talmor2019commonsenseqa}, MedQA \cite{jin2020disease} and OBQA \cite{mihaylov2018suit}. They provide rich scenarios that allow us to thoroughly evaluate the methods' confidence estimation effectiveness in various specialized fields. Detailed information about datasets is shown in the Appendix \ref{appendix:dataset}.

\textbf{Metrics} we adopt two evaluation metrics: AUROC (Area Under the Receiver Operating Characteristic curve) \cite{hendrycks2018baseline,xiong2023llms} and ECE (Expected Calibration Error) \cite{guo2017calibration}, to comprehensively assess the performance of the method. \textbf{AUROC} is a metric for assessing model discrimination between classes. Correct model predictions are marked \textit{positive}, incorrect ones \textit{negative}. The AUROC score spans 0 to 1, with values closer to 1 indicating better performance. 
\textbf{ECE} quantifies how well the model's predicted probabilities are calibrated, meaning the consistency between predicted probabilities and actual occurrence rates. A lower ECE value suggests that the model's predicted probabilities align more closely with the actual outcomes.
The specific calculation details are introduced in the Appendix \ref{appendix:metric}.

\textbf{Baselines}
We compare our method with the following strong baselines:
\begin{itemize}
  \item Verb \cite{lin2022teaching,tian2023just}. It prompts the LLM to assess its confidence in its answer. By designing the prompt template, it requires LLM to return the answer's confidence score, ranging from 0 to 1.
  
  \item Self-Consistency \cite{wang2023selfconsistency,xiong2023llms}.
  It estimates confidence by measuring the consistency among multiple candidate outputs generated by the model. It prompts the model to produce several response candidates and then calculate the consistency score among these candidates.
  
  \item Verb \& Surrogate \cite{shrivastava2023llamas}. It uses a surrogate model with available probabilities to assess the main model's confidence in a given question, and then takes a weighted average with the Verb scores.
    
  \item SC \& Surrogate \cite{shrivastava2023llamas}. Similar with Verb \& Surrogate, it calculates a weighted average of the surrogate model probabilities and SC scores.
    
\end{itemize}
More details are introduced in Appendix \ref{appendix:baseline}
\textbf{Model} 
we focus on the confidence estimation of the closed-source model: GPT-4 \cite{openai2023gpt4} which is the strongest large-scale natural language model with advanced text generation and comprehension capabilities.

As complementary models, we select models including GPT-3.5\footnote{https://openai.com/blog/ChatGPT}, Llama2 \cite{touvron2023llama}, QWen-7B \cite{bai2023qwen}, ChatGLM3 \cite{du2022glm} and DeepSeek \cite{deepseekai2024deepseek}. Although these models have fewer parameters or weaker capabilities compared to GPT-4, they provide effective language processing solutions in scenarios with limited resources.

\subsection{Main Results}

\begin{table*}[ht]
\centering
\resizebox{1\textwidth}{!}{
\begin{tabular}{l|ll|ll|ll|ll}
\hline

\multicolumn{1}{c|}{\multirow{2}{*}{Method}} & \multicolumn{2}{c|}{Chemistry}  & \multicolumn{2}{c|}{Computer\_Security} & \multicolumn{2}{c|}{Business\_Ethics} & \multicolumn{2}{c}{Anatomy} \\
\cline{2-9} 

\multicolumn{1}{c|}{} & AUROC ↑ & \multicolumn{1}{c|}{ECE ↓} & AUROC ↑ & \multicolumn{1}{c|}{ECE ↓} & AUROC ↑ & \multicolumn{1}{c|}{ECE ↓} & AUROC ↑ & ECE ↓  \\ \hline\hline

Verb & 0.682 & \multicolumn{1}{l|}{0.240}       & 0.781        & \multicolumn{1}{l|}{0.138} & 0.746 & \multicolumn{1}{l|}{0.088}                            & 0.642 & \multicolumn{1}{l}{0.145} \\  

Verb \& Surrogate  & 0.699         & \multicolumn{1}{l|}{0.210}       & 0.826        & \multicolumn{1}{l|}{0.092}       & 0.837 & \multicolumn{1}{l|}{0.100} & 0.689 & \multicolumn{1}{l}{0.100} \\   

SC & 0.769 & \multicolumn{1}{l|}{0.180} & 0.787 & \multicolumn{1}{l|}{0.111}  & 0.720 & \multicolumn{1}{l|}{0.080}   & 0.754  & \multicolumn{1}{l}{0.132}   \\ 

SC \& Surrogate & 0.779     & \multicolumn{1}{l|}{0.192}     & 0.797        & \multicolumn{1}{l|}{0.229}       & 0.830        & \multicolumn{1}{l|}{0.075}           & 0.819        & \multicolumn{1}{l}{0.100}    \\ \hline \hline

MPC-Internal     & 0.771       & \multicolumn{1}{l|}{0.189}       & 0.810        & \multicolumn{1}{l|}{0.120}       & 0.892        & \multicolumn{1}{l|}{0.100}     & 0.826        & \multicolumn{1}{l}{0.110}    \\ 

MPC-Across  & 0.783   & \multicolumn{1}{l|}{0.196}    & 0.821    & \multicolumn{1}{l|}{0.091}       & 0.844        & \multicolumn{1}{l|}{\textbf{0.070}}     & {0.834}        & \multicolumn{1}{l}{0.120}   \\ 

MPC   & {\textbf{0.795}}        & \multicolumn{1}{l|}{\textbf{0.151}}      & {\textbf{0.841}}        & \multicolumn{1}{l|}{\textbf{0.075}}    &{\textbf{0.916}}         & \multicolumn{1}{l|}{\textbf{0.070}}                 & {\textbf{0.878}}        & \multicolumn{1}{l}{\textbf{0.009}}    \\ \hline \hline

\multicolumn{1}{c|}{\multirow{2}{*}{Method}}   & \multicolumn{2}{c|}{TruthfulQA} & \multicolumn{2}{c|}{CSQA}     & \multicolumn{2}{c|}{MedQA} & \multicolumn{2}{c}{OBQA} \\

\cline{2-9} 

\multicolumn{1}{c|}{}  & AUROC ↑& \multicolumn{1}{c|}{ECE ↓} & AUROC ↑ & \multicolumn{1}{c|}{ECE ↓} & AUROC ↑ & \multicolumn{1}{c|}{ECE ↓} & AUROC ↑ & ECE ↓ \\ \hline\hline'

Verb    & 0.714       & \multicolumn{1}{l|}{0.076}       & 0.71        & \multicolumn{1}{l|}{0.091}       & 0.669        & \multicolumn{1}{l|}{0.159}                            & 0.776        & \multicolumn{1}{l}{0.032} \\ 

Verb \& Surrogate & 0.751         & \multicolumn{1}{l|}{\textbf{0.042}}       & 0.834        & \multicolumn{1}{l|}{0.049}       & 0.691         & \multicolumn{1}{l|}{0.058}                            & 0.875       & \multicolumn{1}{l}{0.186}  \\ 

SC & 0.804       & \multicolumn{1}{l|}{0.090}       & 0.768        & \multicolumn{1}{l|}{0.110}       & 0.789        & \multicolumn{1}{l|}{0.120}      & 0.813        & \multicolumn{1}{l}{0.034} \\ 

SC \& Surrogate & 0.824       & \multicolumn{1}{l|}{\textbf{0.042}}       & 0.847        & \multicolumn{1}{l|}{0.024}       & 0.775        & \multicolumn{1}{l|}{0.030}        ]               & 0.899        & \multicolumn{1}{l}{0.027} \\ \hline \hline

MPC-Internal & 0.834 & \multicolumn{1}{l|}{0.076} & 0.784 & \multicolumn{1}{l|}{0.100} &0.797 & \multicolumn{1}{l|}{0.100} & 0.804 & \multicolumn{1}{l}{0.031} \\ 

MPC-Across  & 0.830 & \multicolumn{1}{l|}{0.050} & 0.842 & \multicolumn{1}{l|}{0.027}       & 0.808        & \multicolumn{1}{l|}{0.060}  & 0.901 & \multicolumn{1}{l}{0.021} \\ 

MPC & \textbf{0.851} & \multicolumn{1}{l|}{\textbf{0.042}} & \textbf{0.844} & \multicolumn{1}{l|}{\textbf{0.020}} &{\textbf{0.851}} & \multicolumn{1}{l|}{\textbf{0.025}} & \textbf{0.908} & \multicolumn{1}{l}{\textbf{0.020}}      
\\ \hline \hline
\end{tabular}
}
\caption{AUROC and ECE of all confidence methods for GPT4. We compare new methods with strong baselines Verb, SC, Verb\&Surrogate and SC\&Surrogate. Both MPC-Internal and MPC-Across can bring improvements. And MPC achieves results beyond the baselines on all eight datasets. Five average values are taken for each experiment.}
\label{result_main}
\end{table*}

In Table \ref{result_main}, we report the performance of all methods on eight public datasets under two metrics: AUROC and ECE. It is found that:

\textbf{MPC-Internal Enhance Confidence Estimation.}
Compared with SC, MPC-Internal significantly outperforms SC across all eight datasets which indicates that adding verifier perspective to the prompt in the SC method leads to more reliable confidence estimation. Taking MedQA as an example, MPC-Internal improves AUROC by 6.2\% and reduce ECE by 0.095. Similar significant improvements are observed on other datasets as well.

\textbf{MPC-Across Improve Confidence Estimation.} 
By comparing the performance of MPC-Across with SC, We observe that MPC-Across consistently outperforms SC across all eight datasets. On average, MPC-Across exhibit approximately a 5\% increase in AUROC and a decrease of 0.028 in ECE compared to SC. It demonstrates the remarkable superiority of MPC-Across, indicating that the Multi-Perspective from other model can enhance the confidence estimation effectiveness compared with a single model.


\textbf{Fusing MPC-Internal and MPC-Across leads to Best Results.} From the result we can observe that, by fusing MPC-Internal and  MPC-Across, MPC outperformes all baselines across nearly all datasets. We argue that MPC-Internal introducing a self-verification step to increase the verifier's perspective and MPC-Across supplementing the external reasoning perspective by utilizing the reasoning ability of external models are to some extent orthogonal.





\subsection{Effect of MPC on overconfidence}




\label{ana:overconfidence}
\begin{figure}[t]
    \centering
    \resizebox{0.95\linewidth}{!}{
    \includegraphics{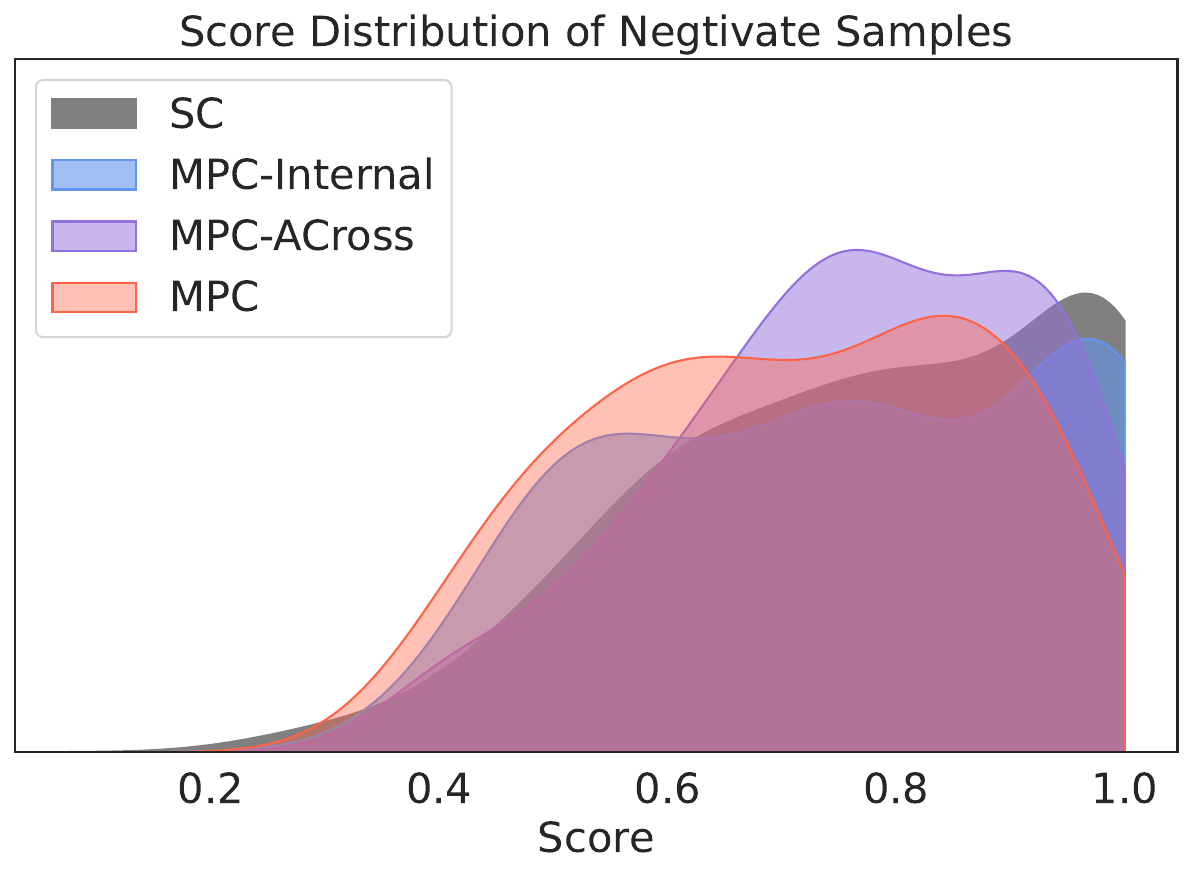}}
    \caption{Confidence score distribution for GPT-4's incorrect samples in TruthfulQA. The horizontal axis represents the predicted confidence. The vertical axis represents the sample density. Note that the distribution shifted more to the left indicates better performance.}
    \label{fig:dis_all}
    \vspace{-0.45cm}
\end{figure}

In order to investigate how MPC mitigates the issue of overconfidence, we analyze the distribution of negative sample scores for GPT-4 on TruthfulQA. The results are shown in Figure \ref{fig:dis_all}. 

\textbf{Effect of MPC-Internal}. 
By comparing the distribution of SC and MPC-Internal, we can observe that applying MPC-Internal reduces the number of overconfident samples with a confidence level between 0.6 and 1.0. And these samples are assigned a lower confidence level ranging from 0.2 to 0.6 indicating MPC-Internal method effectively reduces the confidence level for incorrect answers benefiting from the cross-validation between reasoning and verification perspectives.

\label{overstru}

\textbf{Effect of MPC-Across}. Compared with SC and MPC-Internal, we observe a sharp decrease in the number of negative samples with a high confidence interval of 0.9-1.0 after applying MPC-Across. This significant change indicates that MPC-Across is particularly effective in alleviating overconfidence in high confidence samples with errors.

\textbf{Comparison between MPC-Internal and MPC-Across}. 
MPC-Across significantly alleviate overconfidence in erroneous samples with the confidence score higher than 0.9. These overconfidence stems from the stubborn bias of the main model. In these cases, only by introducing an external model perspective can we effectively reduce the model's overconfidence. On the other hand, MPC-Internal can to some extent alleviate the problem of overconfidence within different score ranges. It reflects that two methods of supplementing perspectives alleviate different overconfidence problems.

\textbf{Effect of MPC}. 
MPC, applying both MPC-Internal and MPC-Across not only reduces overconfidence caused by inherent biases in the main model, but also reduces overconfidence in a wider range of situations, providing a more comprehensive confidence score. The effectiveness of this method is visually demonstrated in Figure \ref{fig:dis_all}, indicating that our proposed MPC has significant potential in improving the accuracy of model prediction confidence.

\subsection{Scalability of MPC}
To verify the effectiveness of our method rather than attributing it to the mere selection of two specific models, we extended our approach to other open-source and closed-source models. The results indicate that MPC can always be effective regardless of model substitution, and even further increasing the number of models can continue to improve confidence estimation. The degree of improvement in MPC varies depending on different parameters such as model structure and size.

\subsubsection{MPC-Internal on GPT-3.5}
We apply MPC-Internal to GPT-3.5 and observe the performance changes it brought. As shown in Table \ref{tab:gpt3.5}, MPC-Internal significantly improve the performance . For example, in Chemistry, MPC-Internal results in a 6.8\% increase in AUROC. Besides, although MPC-Internal slightly reduces AUROC on Business\_ethics, it increases ECE by 0.061. Comparing GPT-3.5 and GPT-4, we find the impact of MPC-Internal on GPT-3.5 is more variable. This may be due to the requirement of MPC-Internal for the verification ability of the main model. Compared to GPT-4, GPT-3.5 has weaker verification capabilities.

\begin{table}[t]
\centering
\resizebox{0.5\textwidth}{!}{%
\begin{tabular}{c|c|c|c|c}
\hline
\multicolumn{1}{c|}{\multirow{2}{*}{Method}}      & \multicolumn{2}{c|}{Chemistry} & \multicolumn{2}{c}{Computer\_security} \\ \cline{2-5} 
    & AUROC ↑    & ECE ↓    & AUROC ↑    & ECE ↓    \\ \hline
SC      & 0.752     & 0.200     & 0.856     & 0.103     \\ 
MPC-Internal      & \textbf{0.820}     & 0.230     & \textbf{0.858}     & \textbf{0.013}   \\ \hline\hline
\multicolumn{1}{c|}{\multirow{2}{*}{Method}}  & \multicolumn{2}{c|}{Business\_ethics} & \multicolumn{2}{c}{Anatomy} \\ \cline{2-5}
      & AUROC ↑    & ECE ↓    & AUROC ↑    & ECE ↓ \\ \hline
SC      &   \textbf{0.835}   & 0.179     & 0.778     & 0.139     \\ 
MPC-Internal      & 0.807     & \textbf{0.118}     & \textbf{0.830}     & \textbf{0.129}     \\ \hline\hline
\end{tabular}%
}
\caption{The performance of MPC-Internal when the main model is GPT-3.5.}
\label{tab:gpt3.5}
\end{table}

\subsubsection{MPC-Across on Other Complementary Models}
In addition to Llama2-70b, we also conduct experiments on the other complementary models: GPT-3.5\footnote{https://openai.com/blog/ChatGPT}, QWen-7B \cite{bai2023qwen}, ChatGLM3-6B-base \cite{du2022glm}, DeepSeek-llm-7B-base \cite{deepseekai2024deepseek} and Llama2-13B \cite{touvron2023llama} to explore the universality of MPC-Across. The confidence score of GPT-3.5 is obtained through SC, while the scores of other open source models are obtained through of token-level probability. Figure \ref{tab:scale} reports the results. 

We find that compared to SC, MPC-Across can improve its performance by using all five complementary models. The improvement brought by different complementary models varies, among which GPT-3.5 has the greatest improvement. For example, on Anatomy, it has brought 12\% and 0.022 improvements to AUROC and ECE, respectively. Results reflect the universality of MPC-Across.

\begin{table}[h]
\centering
\resizebox{0.5\textwidth}{!}{%
\begin{tabular}{c|c|c|c|c}
\hline
\multicolumn{1}{c|}{\multirow{2}{*}{Method}}      & \multicolumn{2}{c|}{Chemistry} & \multicolumn{2}{c}{Computer\_Security} \\ \cline{2-5} 
    & AUROC ↑     & ECE ↓     & AUROC ↑     & ECE ↓      \\ \hline
SC     &0.769	&0.18	&0.787	&0.111     \\ \hline
GPT-3.5 & \textbf{0.842} & \textbf{0.15}& \textbf{0.875}&0.11 \\
Qwen-7B     &0.771	&0.17	&0.845	&0.09      \\ 
ChatGLM3-6B     &0.781	&0.17	&0.828	&0.1    \\
DeepSeek-7B    &0.754	&0.17	&0.835	&0.12   \\
Llama2-13B   &0.779	 &0.19	&0.819	&\textbf{0.08}    \\\hline\hline
\multicolumn{1}{c|}{\multirow{2}{*}{Method}}  & \multicolumn{2}{c|}{Business\_Ethics} & \multicolumn{2}{c}{Anatomy} \\ \cline{2-5}
      & AUROC ↑     & ECE ↓    & AUROC ↑     & ECE ↓  \\ \hline
SC    & 0.72	&0.08	&0.754	&0.132  \\ \hline
GPT-3.5 & \textbf{0.914} & 0.09& \textbf{0.874} & \textbf{0.1} \\
Qwen-7B      &0.894	&\textbf{0.07}	&0.844	&\textbf{0.1}   \\ 
ChatGLM3-6B    & 0.831 &	0.09	&0.833	&0.11    \\
DeepSeek-7B    &0.84	&\textbf{0.07}	&0.842	&0.12\\
Llama2-13B  &  0.796	&0.09	&0.827	&\textbf{0.1}   \\ \hline\hline
\end{tabular}%
}
\caption{Different Complementary Models.}
\label{tab:scale}
\end{table}

\subsubsection{Effect of More Models for MPC-Across}
To verify whether incorporating more additional external models can yield positive effects,we conduct a comparative experiment between two-model and three-model configurations. As shown in Table \ref{tab:num_model}, 2Models refers to the result of weighted average fusion between GPT-4's confidence cores(MPC-Internal) and Llama2-70B's logits-based probabilities. 3Models builds upon 2Models by further incorporating ChatGLM-6B's logits-based probabilities into the weighted average fusion.

The results demonstrate that on four datasets, 3Models consistently outperforms 2Models. This indicates that the reasoning abilities of multiple models can complement each other, leading to superior outcomes. By integrating the diverse perspectives and strengths of more models, we can achieve a more comprehensive confidence estimation.
\begin{table}[t]
\centering
\resizebox{0.5\textwidth}{!}{%
\begin{tabular}{c|c|c|c|c}
\hline
\multicolumn{1}{c|}{\multirow{2}{*}{Method}}      & \multicolumn{2}{c|}{Chemistry} & \multicolumn{2}{c}{Computer\_Security} \\ \cline{2-5} 
    & AUROC ↑    & ECE ↓    & AUROC ↑     & ECE ↓     \\ \hline
MPC-Across &0.78	&0.172	&0.792	&0.12 \\
MPC-2Model &0.836	&0.17	&0.84	&\textbf{0.11}     \\
MPC-3Models &\textbf{0.844}	&\textbf{0.13}	&\textbf{0.862}	&0.12   \\ \hline\hline
\multicolumn{1}{c|}{\multirow{2}{*}{Method}}  & \multicolumn{2}{c|}{Business\_Ethics} & \multicolumn{2}{c}{Anatomy} \\ \cline{2-5}
      & AUROC ↑     & ECE ↓     & AUROC ↑     & ECE ↓  \\ \hline
MPC-Across  &0.844	&0.087	&0.843	&0.09    \\ 
MPC-2Models &\textbf{0.916}	&0.12	&0.877	&0.1 \\
MPC-3Models &0.901	&\textbf{0.07}	&\textbf{0.882}	&\textbf{0.07} \\ \hline\hline
\end{tabular}%
}
\caption{The effect of Multi-Model for MPC.}
\label{tab:num_model}
\end{table}

\subsection{Ablation Study}

\subsubsection{Knowledge Injection}

\begin{figure}
    \centering
    \resizebox{1.0\linewidth}{!}{
    \includegraphics{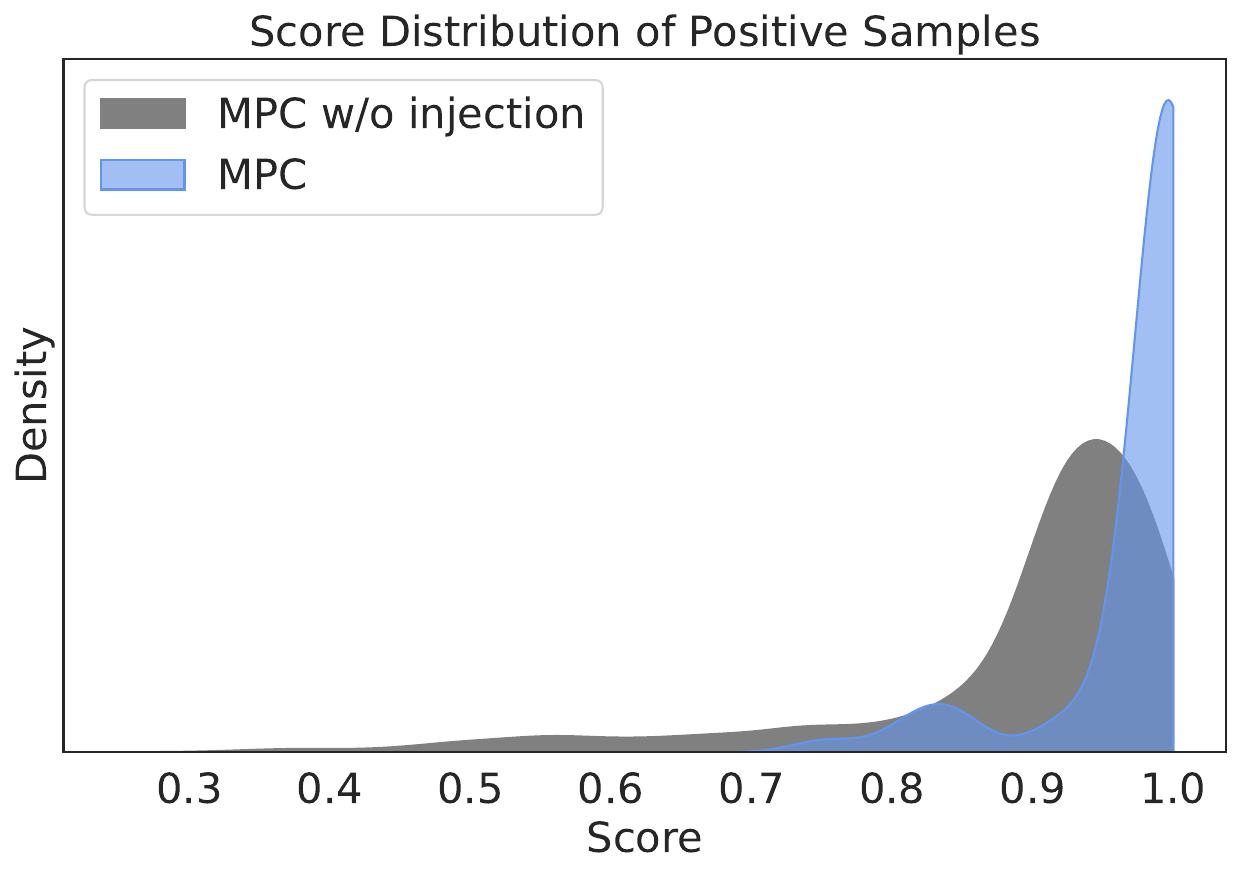}}
    \vspace{-0.5cm}
    \caption{Confidence score distribution for GPT-4's
correct samples in MedQA. The horizontal axis represents the predicted confidence scores. The vertical axis rep-
resents the sample density. Note that the distribution
shifted more to the \textbf{right} indicates better performance.}
    \label{ana:injection}
\end{figure}

In MPC-Across, a key step is to use the main model to generate explanations and provide them to other models. We refer to this step as "knowledge injection", which aims to alleviate confidence estimation errors caused by smaller supplementary models with weak capabilities and lack of necessary reasoning knowledge. In this section, we verify the effectiveness of this step. In Figure \ref{ana:injection}, we demonstrate the impact of injecting or not injecting knowledge on the score distribution of correctly judged samples.

As shown in the Figure \ref{ana:injection}, compared to MPC w/o injection, it is more obvious that the entire distribution shifts to the right, indicating a higher confidence estimate for positive samples. It indicates that "knowledge injection" can enhance the confidence of the supplementary model in the correct answer. It can also be reflected in the ACC of the complementary model. The accuracy of MedQA on Llama2-70b is 48.1\%, while after injecting knowledge, its accuracy increases to 66.2\%. This is also the case on other datasets. Although knowledge injection is not necessary for MPC-Across, it does greatly alleviate the problem of lack of confidence in complementary models on correct samples.

\renewcommand{\dblfloatpagefraction}{.8}

\subsubsection{Robustness of Parameters Alpha}
\label{sec:alpha}

\begin{figure}[t]
    \centering
    \resizebox{1.0\linewidth}{!}{
 \begin{minipage}{0.45\linewidth}
  \vspace{3pt}
  \centerline{\includegraphics[width=\textwidth]{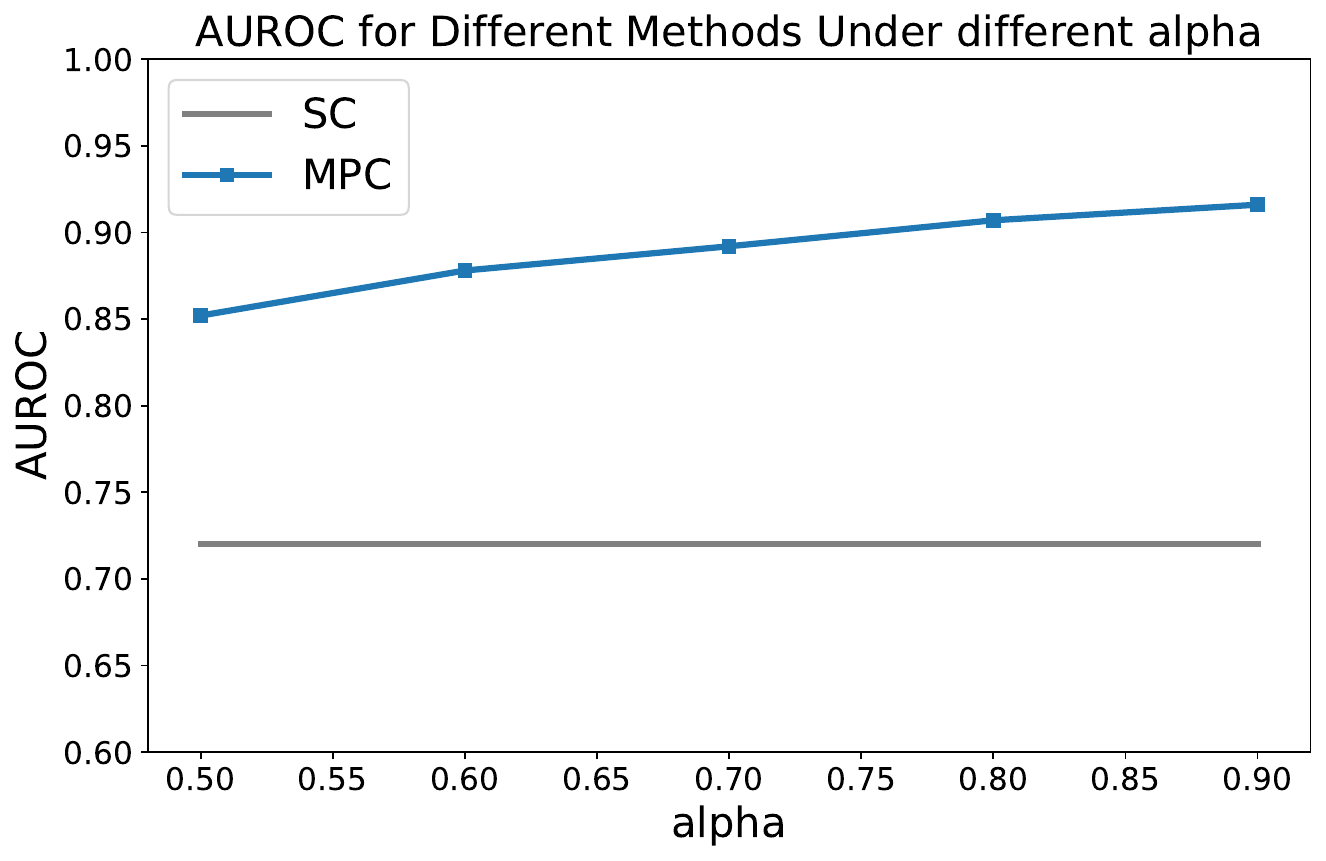}}
  \centerline{(a) Business\_ethics} 
 \end{minipage}
 
 \begin{minipage}{0.45\linewidth}
  \vspace{3pt}
  \centerline{\includegraphics[width=\textwidth]{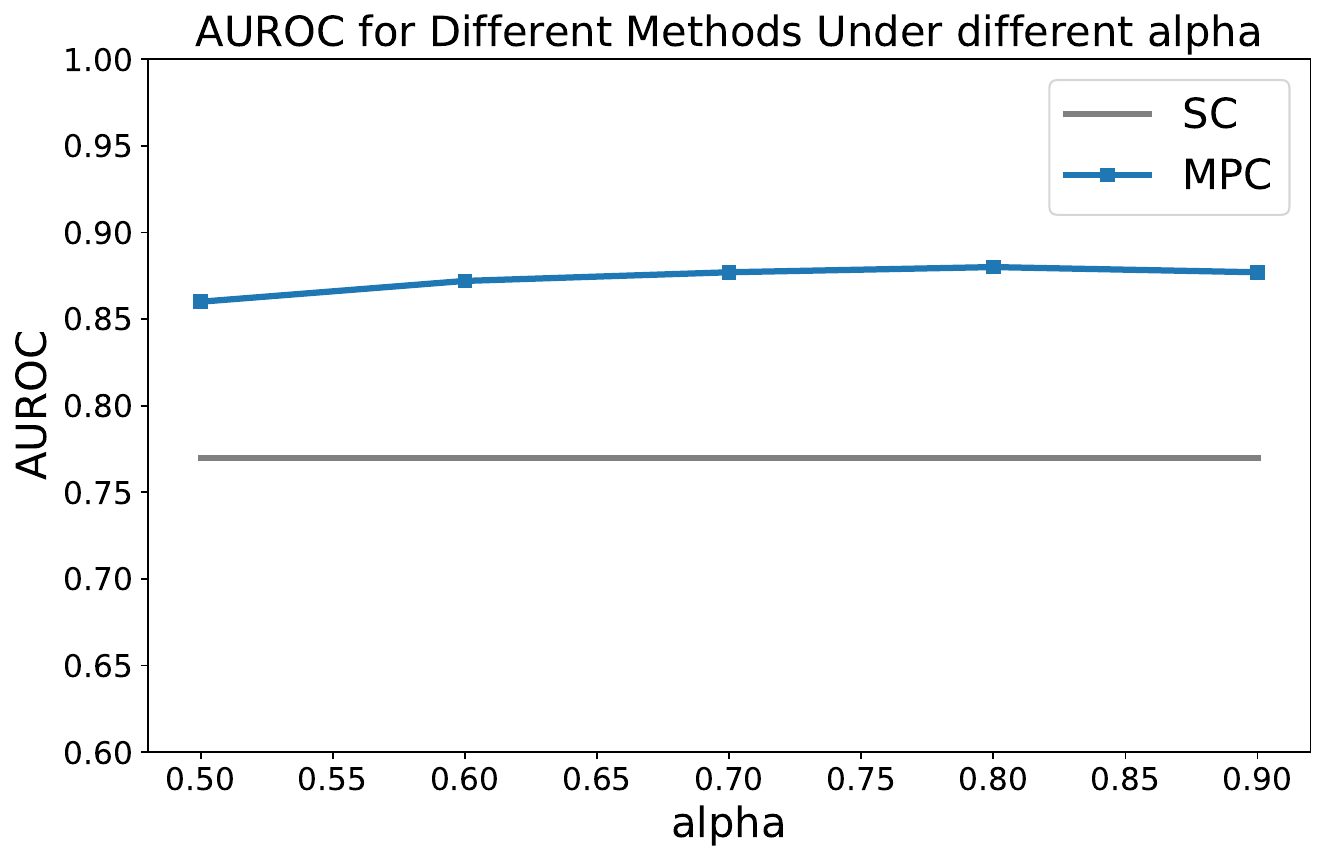}} 
  \centerline{(b) Anatomy}
 \end{minipage}
    }
 \caption{AUROC for MPC under different $\alpha$. We conduct different $\alpha$ values experiments on Business\_Ethics and Anatomy. It shows that our method is robust on different $\alpha$.}
 \label{fig:alpha}
 \vspace{1cm}
\end{figure}

We perform ablation study on the coefficient $\alpha$ to assess model performance across different $\alpha$ values. The results are shown in Figure \ref{fig:alpha}. In the $\alpha$ range of 0.5 to 0.9, MPC far exceeds SC on both datasets. It demonstrates the robustness of our method to hyper-parameter $\alpha$. In addition, by observing the trend of AUROC changes under different $\alpha$ values, we find that the larger the alpha value, the better the performance tends to be. That is to say, when mixing MPC-Internal and MPC-Across, we tend to prefer a larger proportion of MPC-Internal, with MPC-Across as an auxiliary.

\section{Related Work}
\textbf{Confidence Estimation for LLMs} Estimating the confidence level of LLMs' responses is an important research field. \citet{kuhn2023semantic} propose a method called semantic entropy, which incorporates language invariance created by shared meanings to estimate the confidence of LLMs. However, their method requires token-level probability, which is not available for today's closed-source models. \citet{kadavath2022language} guide the model to self-evaluate its answers and directly request LLM to generate the likelihood P of correct answers, while fine-tuning the model to output more accurate likelihood values. \cite{manakul2023selfcheckgpt} propose a sampling based method for detecting hallucinatory facts. All of the above methods involve additional training of models through supervised learning, while recent studies focus on methods that do not require training. \citet{tian2023just} conduct a broad evaluation of computationally feasible methods for extracting confidence scores from LLMs, mainly exploring a large number of variants of verbalized confidence. Meanwhile, \citet{xiong2023llms} investigate three confidence estimation methods for closed-source models, verbalized based, consistency based, and their hybrid methods. \citet{shrivastava2023llamas} propose using the probability of a surrogate model's confidence as the closed-source LLM confidence, and propose a combination of verbalized confidence and surrogate model probability. 

\section{Discussion}
In the current research on model confidence estimation, we have observed several critical issues and phenomena that require greater attention from the research community:

1) \textbf{Are the probabilities of decoding different answers equals to the probabilities of the model's answers being correct?} Confidence estimation is used to identify lower confidence in incorrect answers generated by the model, further to prevent negative impacts in practical applications. The current approaches display a serious overconfidence issue by directly using the probabilities of LLMs output as the probabilities of the answers being correct. We believe that a large number of overconfident samples on incorrect answers precisely demonstrate that these two probabilities are not directly equal, requiring a more in-depth investigation into the form of confidence estimation.

2) \textbf{Lack of Error Self-Awareness.} In this work, we ponder on the question of why we cannot use the probability of decoding answers from models as confidence probabilities. We believe it is due to the model's lack of self-awareness of its own errors. The model is unable to recognize its own mistakes, resulting in it only outputting its own confidence level without considering the accuracy of the answer. This leads to the model giving a high level of confidence in incorrect answers.

3) \textbf{How to alleviate the problem of overconfidence in the model's wrong answers?} In this work, we believe that models are difficult to recognize their own mistakes. Therefore, we propose the MPC method. By collecting the model's answers to the same question from different perspectives, we implicitly make the model aware of its own errors and assign lower confidence scores to overly confident samples. Specifically, we attempt to use the same set of powerful models to generate inconsistent answers to the same question, thereby potentially making it aware of overconfidence issues. In this paper, we refer to this inconsistency as fusion from different perspectives. Regarding the specific classification of perspectives, in this paper, we only brie
fly demonstrate that using different perspectives of the same model itself and different perspectives across models can effectively alleviate overconfidence issues. However, the essence of perspectives, how to add more perspectives, and how to integrate the optimal perspectives are left for future research.

\section{Conclusion}

In this paper, we focus on the issue of overconfidence in confidence estimation for LLMs. We introduce two methods: MPC-Internal and MPC-Across, which alleviate overconfidence through internal self-verification and integration across model perspectives, respectively. Through extensive experiments on multiple datasets, we demonstrate that MPC can effectively improve the accuracy and reliability of confidence estimation. Our work provides a new perspective for improving the confidence estimation of LLMs and lays the foundation for future research.

\section*{Limitation}
In this paper, we introduce the Multi-Perspective Consistency method for confidence estimation of LLMs. Although our method achieves excellent performance, some directions are still to be improved. (1) We focus on that perspective can alleviate overconfidence, but we don't further discuss how to optimally increase the internal perspective. (2) How to combine external models to provide optimal confidence estimates remains to be explored. (3) Our research only involves token-level confidence estimation, and sequence-level confidence estimates remains to be explored. (4) The existing methods obtain confidence after inference, and we believe that obtaining confidence during inference will be a key focus of future research.

\section*{Broader Impact}
Similar to other works on LLM confidence estimation, our research focus is on improving the model's confidence. However, it is important to note that the model itself may generate toxic, harmful and misleading content. In this work, we do not discuss how to address this issue. Future research is needed to explore the ethical and societal implications. It is also important to highlight that our approach is specifically designed for research settings, and its testing has been limited to such environments. It should not be directly applied without further analysis to assess potential harm or bias in the proposed application.

\bibliography{custom}

\appendix

\section{Related Prompt}
We show the prompt used by MPC-Internal in Figure \ref{prompt_psc}. Unlike SC, we add a self-verification step to promote self-reflection of LLM. [one option] can be obtained through a reply from LLM or by randomly matching an option. The prompt of MPC-Across is shown in Figure \ref{prompt_aec}. It shows the process of using the main model to generate knowledge related to the question and enhancing the supplementary model. The relevant knowledge is added to the prompt of the supplementary model to alleviate the insufficient knowledge of the supplementary model.

\label{appendix:prompt_psc}
\begin{figure}[t]
\begin{tcolorbox}[
colback=red!5!white,
colframe=red!60!white,
title=Prompts of SC and PC-Internal]
\textbf{Prompt of SC}

Answer the following question to the best of your ability and give your reason.

Question: [question]

Answer: 

\textbf{Prompt of MPC-Internal}

Answer the following question to the best of your ability.

Question: [question]

Is Answer[one option] the correct one? If so, why? If not, which one is the correct answer? Please reflect on it.

Answer:

\end{tcolorbox}
\caption{Prompts of SC and MPC-Internal}
\label{prompt_psc}
\end{figure}

\begin{figure}[t]
\begin{tcolorbox}[
colback=red!5!white,
colframe=red!60!white,
title=Prompts of MPC-Across]
\textbf{Step 1. Prompt of Main Model}

Generate relevant knowledge for the following question. 

Question: [question]. 

Relevant knowledge:

\textbf{Step 2 Prompt of complementary Model }

Question: [question]

Supplementary Information: [Response of Step 1]

Answer:

\end{tcolorbox}
\caption{Prompts of MPC-Across}
\label{prompt_aec}
\end{figure}

\section{Experiment Settings}
\subsection{Implementation Details}
When using closed-source models GPT-4 and GPT-3.5 for inference, we use the public API provided by Open-AI. For the inference of open-source models, we use greedy decoding strategy. For MPC-Internal, we choose $K=15$, $\alpha=0.8$. We perform each experiment 5 times and report the average results.

\subsection{Dataset}
We utilize eight commonly used public datasets for evaluation. They are: 

\textbf{MMLU} \cite{hendrycks2021measuring} is a benchmark test for evaluating model pre training knowledge, challenging 57 different disciplines with zero and few samples. The test covers a range of basic to professional levels, aiming to identify blind spots in the model's world knowledge and problem-solving abilities. We select four domain subsets including Chemistry, Computer\_Security, Business\_Ethics and Anatomy. 

\textbf{TruthfulQA} \cite{lin2022truthfulqa} is a benchmark designed to measure how authentic language models are at answering questions. The benchmark contains 817 questions across 38 categories, including health, law, finance and politics. The questions in the test are carefully designed to test whether the model gives wrong answers due to false beliefs or misunderstandings. 

\textbf{CSQA}\cite{talmor2019commonsenseqa} is a dataset used to evaluate the ability of AI models to answer commonsense questions, consisting of 12247 multiple-choice questions, requiring the model to use prior knowledge to distinguish subtle conceptual differences. We randomly selected 1000 from the test set as the evaluation set for confidence estimation.

\textbf{MedQA} \cite{jin2020disease} is a multilingual open domain question answering dataset in the medical field, including free form multiple-choice questions. We only select the English subset as the confidence test set.

\textbf{OBQA} \cite{mihaylov2018suit} is a question and answer dataset based on the open book exam model, containing 5957 basic-level science multiple-choice questions to evaluate understanding of core scientific knowledge and its application in new contexts. We select all 500 test sets to test confidence estimation.
We show the test set size of each dataset in Table \ref{dataset}

\begin{table}[h]
\centering
\resizebox{0.48\textwidth}{!}{%
\begin{tabular}{|c|c|c|c|}
\hline
Dataset & Test set size & Dataset & Test set size \\ \hline
TruthfulQA & 817 & Chemistry&100 \\ \hline
CSQA & 1000 & Computer\_Security& 100\\ \hline
MedQA & 1273 & Business\_Ethics&  100\\ \hline
OBQA & 500 &  Anatomy &135 \\ \hline

\end{tabular}
}
\caption{Statistics of Datasets.}
\vspace{-0.5cm}
\label{dataset}
\end{table}

\label{appendix:dataset}
\subsection{Metrics}
\label{appendix:metric}
\textbf{AUROC} (Area Under the Receiver Operating Characteristic Curve) is a common metric used to measure the classifier's discriminative ability.  We define the function R (x, y) to indicate that for a given input x, if the predicted answer y is correct, then R (x, y) is 1, otherwise it is 0. Meanwhile, C (x) represents the model's confidence in predicting sample x, with a value between 0 and 1.
True Positive Rate (TPR) is defined at the confidence threshold t as the proportion of correctly predicted samples with a confidence level not lower than t, and its calculation formula is:

\vspace{0.5mm} 
\[ \text{TPR}(t) = \frac{\mathbb{E}[R(x, y(x)) \cdot \mathbb{I}(C(x) \ge t)]}{\mathbb{E}[R(x, y(x))]} \]
\vspace{0.5mm} 

False Positive Rate (FPR) is defined as the proportion of incorrectly predicted samples with a confidence level of no less than t at the confidence threshold t. Its calculation formula is:


\vspace{0.5mm} 
\[
\text{FPR}(t) = \frac{\mathbb{E}[(1 - R(x, y(x))) \cdot \mathbb{I}(C(x) \ge t)]}{\mathbb{E}[1 - R(x, y(x))]}
\]
\vspace{0.5mm} 

Draw TPR(t) and FPR(t) values at different thresholds t to form ROC curves. Then, calculate the area under the ROC curve, which is AUROC. This area reflects the ability of the model to classify correctly based on the threshold.

\textbf{ECE} (Expected Calibration Error) is a metric that quantifies the level of model calibration. An ideal confidence estimation method should reflect the probability of correctness. ECE calculates the calibration error of a model by comparing its predicted probability with the actual frequency of occurrence. Specifically, it is obtained by dividing the predicted probability into several intervals (bin), and then calculating the weighted average of the difference between the average predicted probability and the actual occurrence frequency within each interval.

\vspace{0.01mm}
\[
 \text{ECE} = \sum_{m=1}^M \frac{|B_m|}{n} | \text{acc}(B_m) - \text{conf}(B_m) | 
\]
\vspace{0.01mm} 

$M$ is the number of the bins. $B_m$ is the sample set within the $m_{th}$ bin. $n$ is the total number of samples. $acc(B_m)$ is the accuracy of the samples within $B_m$, which is the proportion of correct predictions for these samples. $\text{conf}(B_m)$ is the average confidence score of the samples within ${B_m}$.

\subsection{Baselines}
\label{appendix:baseline}
We compare our method with the following strong baselines:
\begin{itemize}
\item Verb \cite{lin2022teaching,tian2023just}. It prompts the LLM to assess its confidence in its answer. By designing the prompt template, it requires LLM to return the answer and its confidence score for the answer, ranging from 0 to 1. In addition, the model will be required to generate the COT process. Due to the fragility of self-awareness in LLMs,the scores are mostly concentrated between 0.8-1.0, indicating a serious problem of overconfidence.
  
  \item Self-Consistency \cite{wang2023selfconsistency,xiong2023llms}.
  It estimates confidence by measuring the consistency among multiple candidate outputs generated by the model. It prompts the model to produce several response candidates and then calculate the consistency score among these candidates. For closed-source models where logits are not availabel, the scores obtained by the self-consistency can to some extent serve as a substitute for logits. Our analysis experiment indicates that it also leads to overconfidence.
  
  \item Verb \& Surrogate \cite{shrivastava2023llamas}. It believes that the logits-based score of the open-source model can be used as the confidence score of the closed-source model. Therefore, it uses a surrogate model with available probabilities to assess the answers of the main model in a given question, and then takes a weighted average with the Verb scores.
    
  \item SC \& Surrogate \cite{shrivastava2023llamas}. Similar with Verb \& Surrogate, it calculates a weighted average of the surrogate model probabilities and SC scores. Unlike it, our MPC-Across is not limited to the open-source model, but can be extended to any model. Besides, we add a process of injecting knowledge to alleviate the impact of weak surrogate model capabilities.
    
\end{itemize}

\section{Different K values of SC}

\label{sec:k}
\begin{figure}
    \centering
    \resizebox{1.0\linewidth}{!}{
    \includegraphics{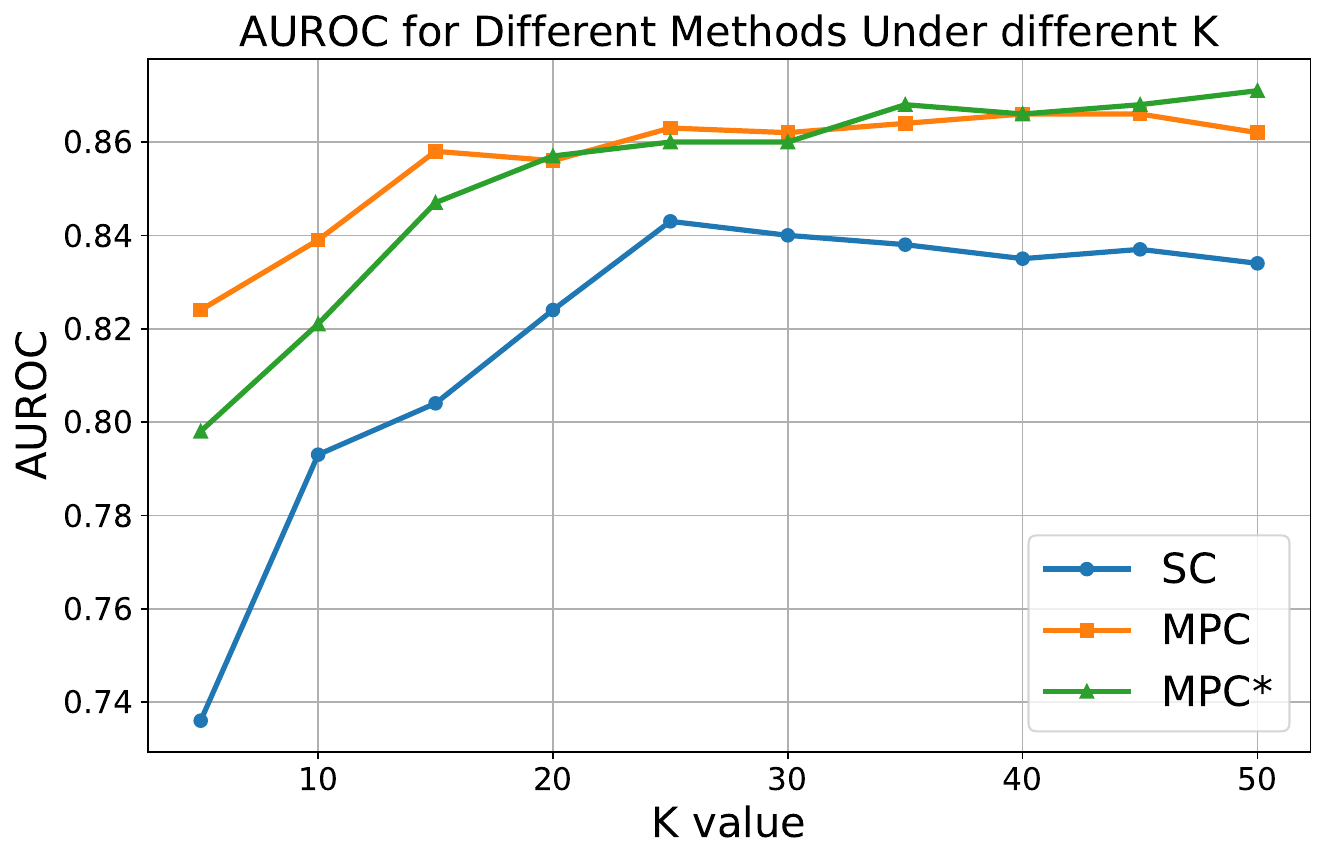}}
    \caption{AUROC for SC, MPC, MPC* under different K on TruthfulQA. For MPC*, we abandon the initial step of generating answers using LLM. Instead, answer is matched randomly for the model's self-reflection.}
    \label{fig:K}
    \vspace{-0.45cm}
\end{figure}

To validate the effectiveness of MPC-Across on various K values, we conduct experiments ranging from K=5 to K=50. K means the number of answers when conducting MPC-Internal. As shown in Fig \ref{fig:K}, \textbf{it shows a consistent positive impact on SC throughout all K values}, with an average improvement of about 2\%.

\renewcommand{\dblfloatpagefraction}{.9}
Besides, we introduce a variant of MPC, \textbf{MPC*}. The difference is for MPC*, we abandon the initial step of generating the answer for self-reflection. Instead, we random select an answer for self-reflection in each step. MPC* requires greater verification ability. Comparing MPC with MPC*, MPC performs better at lower K. However, with K values above 20, MPC* can achieve results equal to or even exceed MPC. This may be because the importance of generating high-quality and accurate answers becomes more significant when the number of responses is low. However, as the K value gradually increases, at higher iterations, the strategy of randomly selecting answer validation can give the model more opportunities to explore and verify different answers, thus to some extent compensating for the model's lack of self-validation ability.

\end{document}